\documentclass[lettersize,journal]{IEEEtran}
\usepackage{amsfonts}
\usepackage{algorithmic}
\usepackage{algorithm}
\usepackage{array}
\usepackage[caption=false,font=normalsize,labelfont=sf,textfont=sf]{subfig}
\usepackage{textcomp}
\usepackage{stfloats}
\usepackage{url}
\usepackage{verbatim}
\usepackage{graphicx}
\usepackage{cite}
\usepackage{booktabs}
\usepackage{multirow}
\usepackage{amsmath}
\usepackage{hyperref}
\usepackage{multirow}
\usepackage{tabularx}
\usepackage{booktabs}
\usepackage{multirow}
\usepackage[table]{xcolor}
\usepackage[table,xcdraw]{xcolor}
\usepackage[normalem]{ulem}
\useunder{\uline}{\ul}{}
\allowdisplaybreaks
\usepackage[table,xcdraw]{xcolor}
\usepackage{amssymb}
\usepackage{amsmath}
\usepackage{booktabs}
\usepackage{multirow}
\usepackage{paralist,bm}
\usepackage{tabularx}    
\usepackage{booktabs}    
\usepackage{multirow}    
\usepackage{graphicx}    

\begin{document}

\title{RadioFormer3D: Weakly Supervised 3D Radio Map Estimation in Low-Altitude Airspace via Generative Modeling}

\author{Zheng Fang, Junjie Liu, Kangjun Liu, Jianguo Zhang, Yaowei Wang~and~Ke Chen,~\IEEEmembership{Member, IEEE} 
\thanks{This work is supported in part by the Major Key Project of Pengcheng Laboratory under Grant No. PCL2025A14 and PCL2025A02, and the National Natural Science Foundation of China under Grant No. 62536003.
\textit{(Corresponding authors: Ke Chen and Kangjun Liu.)}}

\thanks{Z. Fang and J. Liu are with the Pengcheng Laboratory, China, and also with the Department of Computer Science and Engineering, Southern University of Science and Technology, Shenzhen, China~(emails:fangzh01@pcl.ac.cn; junjieliu$\_$cn@outlook.com);
K. Liu and K. Chen are with the Pengcheng Laboratory, China~(emails:liukj@pcl.ac.cn; chenk02@pcl.ac.cn); 
J. Zhang is with the Department of Computer Science and Engineering, Southern University of Science and Technology, Shenzhen, China~(email: zhangjg@sustech.edu.cn);
Y. Wang is with the Harbin Institute of Technology, Shenzhen, China, and also with the Pengcheng Laboratory, China~(email: wangyaowei@hit.edu.cn).
}
}

\markboth{Journal of \LaTeX\ Class Files,~Vol.~14, No.~8, August~2021}%
{Shell \MakeLowercase{\textit{et al.}}: A Sample Article Using IEEEtran.cls for IEEE Journals}


\maketitle
\begin{abstract}
With the emergence of wireless applications in three-dimensional environments, such as the low-altitude airspace and 3D heterogeneous networks, radio map estimation is increasingly required to characterize signal propagation across both horizontal and vertical dimensions. 
However, extending radio map estimation from 2D to 3D remains challenging due to increased spatial sparsity and limited supervision across continuous altitudes. 
In this paper, we propose \textbf{\textit{RadioFormer3D}}, a specialized model for volumetric spectrum reconstruction under weak supervision. 
Building on the dual-stream, multi-granularity fusion architecture of \textit{RadioFormer}, \textit{RadioFormer3D} introduces a Fourier-based sampling encoder and a volumetric decoder to efficiently process sparse measurements in 3D space. 
To alleviate the lack of vertical supervision, we propose the \textbf{\textit{Joint Spectrum Integrity Loss}}, which integrates volume-level pseudo-label supervision, map-level geometry-aware radio rendering, and pixel-level localized constraints within a unified optimization scheme. 
This design enables the model to capture complex vertical structural relationships more effectively under sparse supervision. 
Extensive experiments across several radio map datasets show that \textit{RadioFormer3D} achieves superior overall performance compared to representative existing methods. 
In particular, it demonstrates improved reconstruction quality at unlabeled altitudes while maintaining a favorable trade-off between accuracy and inference efficiency,
positioning it as a highly promising solution for future 3D environment-aware wireless networks.
\end{abstract}


\section{Introduction}

\IEEEPARstart{T}{he} proliferation of next-generation wireless networks, particularly the vision of 6G and Integrated Sensing and Communication, as well as the rapid emergence of low-altitude economy applications, has transformed the electromagnetic spectrum into a highly dynamic and multidimensional resource~\cite{6g, 6g2}. 
Within this context, Electromagnetic Spectrum Situation Awareness has emerged as a critical capability, providing the information required for autonomous network optimization, cognitive interference mitigation, and the management of reconfigurable intelligent surfaces~\cite{essa, ris}. 
However, due to hardware costs and deployment constraints, obtaining a real-time, high-fidelity spectrum view remains a significant challenge. To address this limitation, spectrum situation generation or radio map estimation~(RME) algorithms, which reconstruct a continuous, dense global radio map from spatially sparse measurements, have become an indispensable bridge to maximize spectrum efficiency and ensure network reliability~\cite{radiounet, 3dsparse}.

\begin{figure}[t!]
\centering
\includegraphics[width=3.4 in]{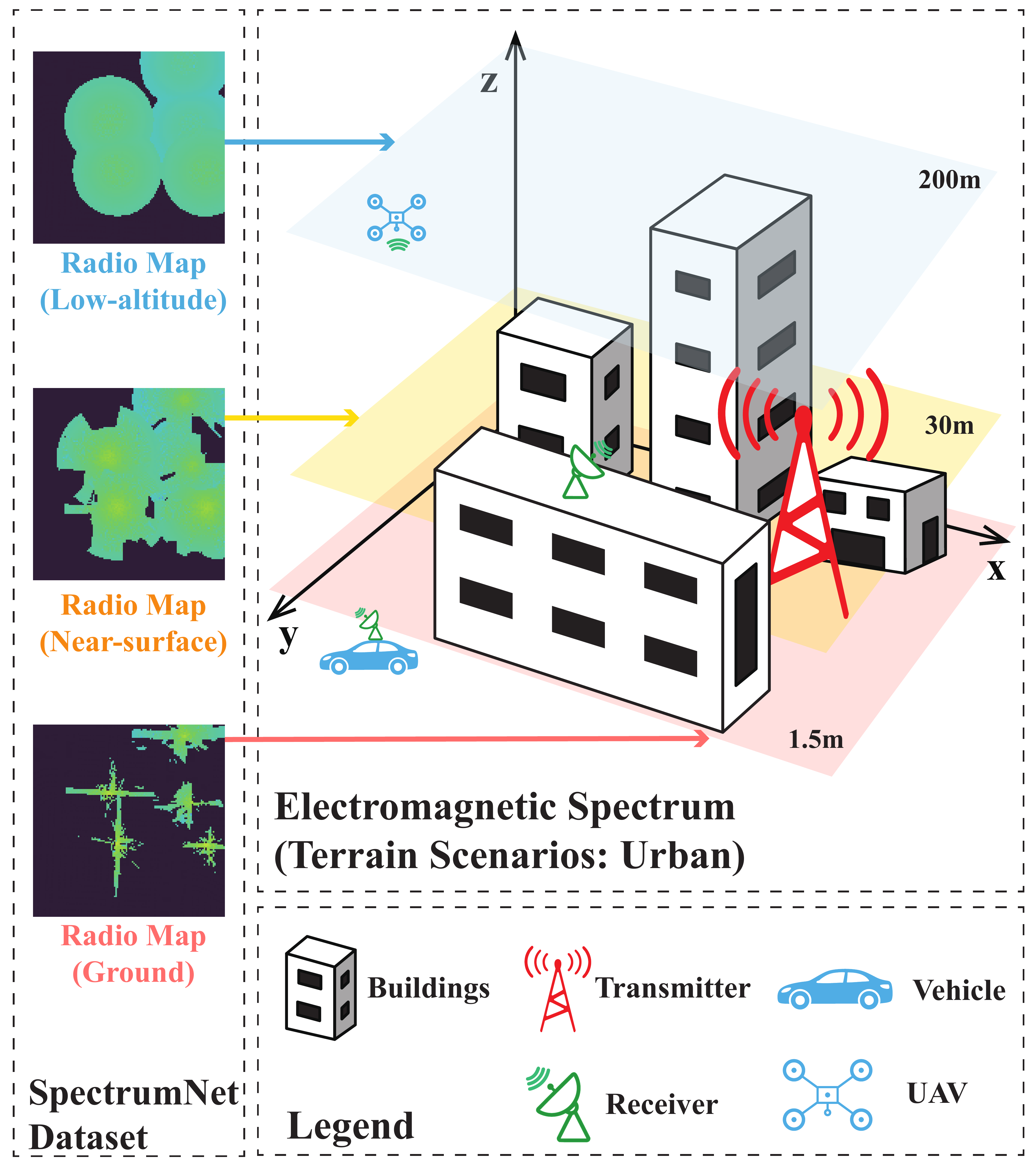}
\caption{Illustration of the 3D electromagnetic spectrum situation in urban scenarios. 
The left section displays the hierarchical radio maps from the SpectrumNet dataset at three representative altitudes: Ground~(1.5m), Near-surface~(30m), and Low-altitude~(200m). 
The right section depicts the corresponding physical environment, highlighting the interactions among electromagnetic propagation, building geometries, and heterogeneous receivers~(e.g., UAVs and vehicles).}
\label{fig_1}
\end{figure}

Traditionally, spectrum situation awareness has characterized the electromagnetic environment primarily through 2D planar images~\cite{radiomap}, which represent signal distributions at specific, fixed altitudes. 
Driven by the rapid expansion of wireless services into the vertical dimension, the transition from 2D image-level modeling to 3D volumetric characterization is now necessitated.
Since traditional planar maps fail to capture the complex spatial interactions required by emerging sectors, the rapid expansion of UAV-assisted and 3D heterogeneous networks has shifted the focus of spectrum awareness from ground-level connectivity to holistic volumetric space~\cite{space, new}.
As illustrated in Fig.~\ref{fig_1}, this shift toward volumetric characterization stems from the critical realization that electromagnetic environments exhibit profound vertical heterogeneity, a complexity that traditional 2D models fail to reconcile. 
At ground level, the radio map is severely fragmented by dense urban fabric, where building shadowing and multi-path reflections induce highly non-linear signal variations. 
In stark contrast, as altitude increases toward the low-altitude airspace, the signal distribution transcends terrestrial obstructions and increasingly approximates free-space propagation. 
Such drastic physical divergence along the altitude axis renders conventional 2D planar modeling insufficient to capture the volumetric height of contemporary connectivity. 
Accordingly, to effectively leverage multi-tiered observation systems that incorporate both terrestrial sensors and UAV-mounted collectors, it is imperative to develop a new generation of algorithms capable of high-fidelity 3D spectral reconstruction~\cite{spasebaysian, multi}.

Recent years have witnessed increasing interest in the RME task, with many studies focusing on improving reconstruction performance by developing more effective network architectures. 
Among them, \textit{RadioFormer}~\cite{radioformer} stands out as a representative approach due to its dual-stream multi-granularity fusion design, which achieves a favorable balance between prediction accuracy and inference efficiency. 
Its architecture provides an effective paradigm for exploiting sparse radio measurements by combining complementary structural and local representations. 
Nevertheless, as RME moves from 2D planar settings toward 3D volumetric environments, existing methods, including \textit{RadioFormer}, face several new difficulties. 
The first challenge arises because the transition from 2D planes to 3D manifolds exacerbates the signal's inherent sparsity. 
As the observation space scales, limited measurements become increasingly sparse in information density throughout the volume. 
This geometric expansion significantly amplifies the difficulty of extracting meaningful spatial correlations. 
In that context, maintaining global consistency under minimal local guidance requires superior feature aggregation. 
Furthermore, it requires the models to bridge the vertical gap while mitigating the intensified sparsity of spatial signals~\cite{bayactive, recon}.
The second challenge involves a vacuum in vertical supervision. 
Unlike 2D datasets where dense ground truth is readily available~\cite{radiomap}, constructing a 3D spectrum dataset is an extraordinarily resource-intensive endeavor. 
It requires high-fidelity ray tracing across massive voxel volumes. 
As a result, existing benchmarks typically provide only a few discrete altitude samples. 
This leaves the majority of the 3D space spectrally latent during training~\cite{weaksup}. This lack of labels forces models to move beyond simple data fitting to infer complex spatial coupling across heterogeneous altitudes. 
Without dense vertical supervision, capturing the intricate signal transitions between ground-level clutter and low-altitude free space often leads to representation collapse in conventional architectures.

To address these challenges, we propose \textbf{\textit{RadioFormer3D}}, a generative model designed for 3D radio map estimation. 
While retaining \textit{RadioFormer}'s multi-granularity fusion architecture~\cite{radioformer}, we implement structural enhancements to adapt the network to volumetric manifolds. 
Specifically, a Fourier-based sampling encoder maps sparse measurements into high-dimensional space to ensure global geometric consistency. 
Additionally, a specialized volumetric decoder reconstructs complex signal distributions over continuous altitude, enhancing the model’s ability to handle extreme sparsity efficiently.
To bridge the vertical supervision gap, we introduce the \textbf{\textit{Joint Spectrum Integrity Loss}}, which provides consistent physical guidance across heterogeneous altitudes. 
This multi-level strategy integrates three complementary granularities: volume-level linear pseudo-labeling to infer global signal trends at unobserved altitudes, map-level radio rendering to enforce structural consistency between signal fields and physical obstacles, and pixel-level constraints to anchor the model to original sparse measurements. 
By blending these granularities, the model effectively captures complex vertical wave coupling without requiring dense 3D ground truth.
The primary contributions of this work are summarized as follows:
\begin{itemize}
\item We study 3D radio map estimation under weak supervision and propose \textbf{\textit{RadioFormer3D}} as a dedicated model for this task.
By extending the multi-granularity fusion architecture with a Fourier-based encoder and a volumetric decoder, our model effectively addresses the challenges posed by signal sparsity in 3D radio environments.
\item We introduce the \textbf{\textit{Joint Spectrum Integrity Loss}} as a multi-level supervision strategy to alleviate the lack of vertical labels. 
It combines volume-level pseudo-labeling, geometry-aware map rendering, and pixel-wise regression to encourage structural consistency in the reconstructed fields while remaining faithful to sparse observations.
\item We conduct comprehensive experiments on the \textit{UrbanRadio3D} and \textit{SpectrumNet} datasets to evaluate the proposed method. 
The results show that \textit{RadioFormer3D} achieves strong performance in both reconstruction accuracy and generalization, while the \textit{Joint Spectrum Integrity Loss} further brings consistent improvements to various backbones under weak supervision.
\end{itemize}

The remainder of this paper is organized as follows: 
Sec.~\ref{sec:related_work} reviews prior studies on radio map estimation, covering classical propagation-based methods, deep learning approaches for supervised radio map prediction, and recent advances in 3D radio field reconstruction.
Sec.~\ref{method} describes the architecture and overall framework of the proposed \textit{RadioFormer3D}. 
Sec.~\ref{exp} presents the experimental settings and reports both quantitative and qualitative results for performance evaluation.
Finally, Sec.~\ref{conclusion} concludes the article with a summary of findings and implications.
The code will be released at \href{https://github.com/FzJun26th/RadioFormer3D}{https://github.com/FzJun26th/RadioFormer3D}.
\section{Related Work}
\label{sec:related_work}

\subsection{Radio Map and Path Loss}
Conventional RME primarily relies on deterministic simulations and empirical propagation models. Ray-tracing~\cite{raytracing1} is the most fundamental deterministic approach for simulating radio propagation, providing high accuracy by modeling electromagnetic wave interactions with building geometries. This method is often grounded in the log-distance path loss model to characterize signal attenuation~\cite{rayt2}:
\begin{equation}
    PL(d) = PL(d_0) + 10n \log_{10}\left(\frac{d}{d_0}\right) + X_{\sigma},
    \label{eq:log}
\end{equation}
where $n$ represents the path loss exponent and $X_{\sigma}$ denotes shadowing effects~\cite{rayt3}. To generate a dense radio map, a ray-tracing engine must iteratively calculate the trajectories of thousands of rays at each grid point, accounting for reflection and diffraction. 
Consequently, its computational cost is several orders of magnitude higher than deep learning-based models, making it prohibitive for real-time applications~\cite{rayt4}. 
This limitation is further exacerbated in 3D scenarios. 
While 2D ray tracing scales with the number of obstacles in a plane, 3D ray tracing must account for vertical diffraction and multi-surface reflections across a volumetric manifold. 
This geometric expansion leads to an exponential increase in the search space for valid ray paths, rendering high-fidelity 3D volumetric simulation extremely resource-intensive.

\subsection{Deep Learning-based Radio Map Estimation}
In recent years, deep learning has emerged as a powerful paradigm for the RME task due to its superior non-linear fitting capabilities. 
Early attempts predominantly used general-purpose architectures such as \textit{UNet}~\cite{unet} for spectrum generation tasks. 
Building on this, \textit{RadioUNet}~\cite{radiounet} was proposed as the first specialized deep learning model for RME, integrating building geometry to predict 2D path loss. 
To further enhance the structural modeling capability of the UNet-based backbones, \textit{DAT-UNet}~\cite{datunet} introduced deformable convolutional networks~\cite{dcnv1,dcnv2} to capture complex spatial dependencies and geometric distortions.
To improve physical consistency, several studies have sought to integrate radio-propagation heuristics into neural networks. 
Notable examples include \textit{RadioDUN}~\cite{radiodun}, which employs deep unfolding techniques to mimic iterative optimization, and \textit{PMNet}~\cite{pmnet}, which incorporates path-loss models directly into its architecture. 
More recently, \textit{RadioFormer}~\cite{radioformer} introduced a multi-granularity Transformer~\cite{transformer} framework to achieve high-fidelity reconstruction while maintaining computational efficiency. 

Regardless of their architectural differences, these models are typically trained using a standard Mean Squared Error~(MSE) loss to minimize the pixel-wise discrepancy between the predicted radio map $\hat{\mathbf{R}}$ and the ground truth $\mathbf{R}$:
\begin{equation}
    \mathcal{L}_{\mathrm{MSE}} = \frac{1}{N_p} \sum (\mathbf{R} - \hat{\mathbf{R}})^2,
    \label{eq:mse_loss}
\end{equation}
where $N_p$ denotes the number of pixels in the radio map. 
Despite their success, these methods are predominantly confined to 2D planar estimation and rely on dense supervision. 
As such, they often fail to address the ``supervision vacuum'' challenge in 3D volumetric scenarios.

\subsection{3D Radio Field Reconstruction} 
Recent advancements in computer vision, particularly Neural Radiance Fields~(NeRF)~\cite{mildenhall2021nerf} and 3D Gaussian Splatting~(3DGS)~\cite{kerbl20233d}, have inspired new paradigms for 3D radio field modeling from sparse signal measurements. 
\textit{NeRF$^{2}$}~\cite{zhao2023nerf2} extends neural radiance fields to radio-frequency propagation and reconstructs a continuous radio radiance field from sparse signal measurements in a fixed scene. 
\textit{NeWRF}~\cite{lu2024newrf} similarly builds a NeRF-based wireless radiation field from sparse channel measurements to predict channels at unobserved locations, while \textit{WRF-GS}~\cite{wen2025neural} leverages the efficiency of 3DGS to achieve high-fidelity radio map extrapolation and real-time reconstruction. 
Despite differences in representations and rendering strategies, these methods are fundamentally built for scene-specific radio field reconstruction, requiring a dedicated training or optimization process for each new environment. 
Such a formulation limits scalability and prevents efficient deployment in unseen scenes. 
In contrast, we study feed-forward reconstruction of the electromagnetic spectrum, aiming to learn scene-agnostic representations that generalize across environments. Thus, our model can directly predict the spectrum field in unseen scenes through a single forward pass, without per-scene fitting.
\section{Method}
\label{method}

This section details the methodology of the proposed \textit{RadioFormer3D} framework. We first formulate the 3D radio map estimation problem under weak supervision. Subsequently, we present the multi-granularity network architecture designed to handle extreme spatial sparsity. Finally, we formulate the Joint Spectrum Integrity Loss, which governs the model optimization to bridge the vertical supervision gap.

\begin{figure*}[t]
\centering
\includegraphics[width=\textwidth]{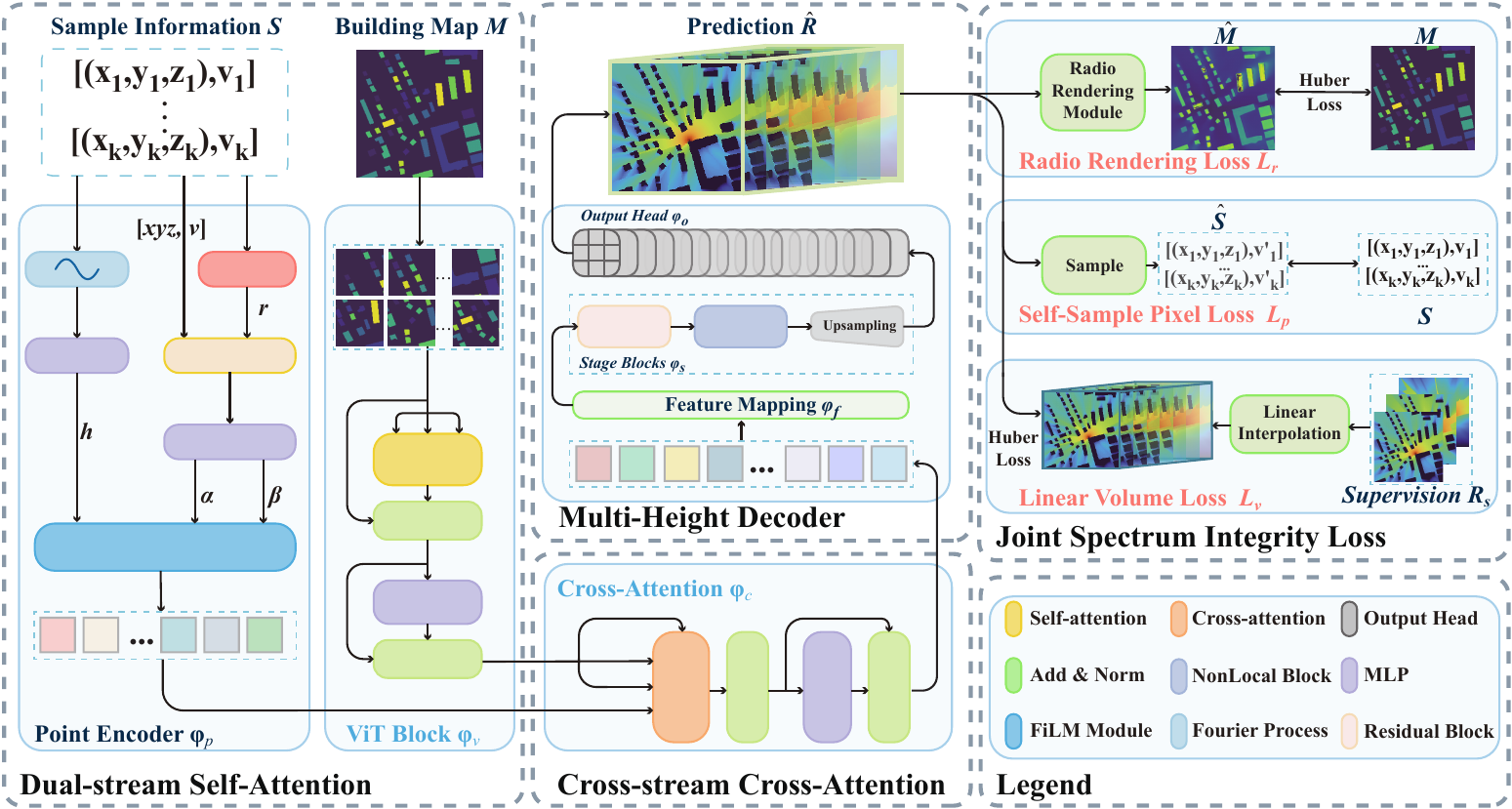}
\caption{Architectural overview of the proposed \textit{RadioFormer3D}. 
The framework adopts a dual-stream design to process heterogeneous inputs for 3D RME. 
The DSA module independently encodes multi-granularity features: the building-height map $\mathbf{M}$ is processed by ViT blocks to capture environmental context, while the sparse sample information $S$ is encoded via a Point Encoder integrated with Fourier and FiLM processes. 
These parallel features are subsequently fused through the CSCA module to capture complex spatial-spectral correlations. 
The resulting fused representation is then reconstructed into a volumetric spectrum space by the Multi-Height Decoder. 
To facilitate robust training under weak supervision, the framework is optimized using the \textit{Joint Spectrum Integrity Loss}, which incorporates three granularities: \textit{Radio Rendering Loss} $\mathcal{L}_{\mathrm{r}}$, \textit{Self-Sample Pixel Loss} $\mathcal{L}_{\mathrm{p}}$, and \textit{Linear Volume Loss} $\mathcal{L}_{\mathrm{v}}$.}
\label{fig_main}
\end{figure*}

\subsection{Problem Formulation}
In this subsection, we formally define the task of 3D radio map estimation (3D RME) and introduce the notations used throughout this paper.

The objective of 3D radio map estimation is to reconstruct a dense, volumetric representation of electromagnetic quantities across a geographic region. Unlike traditional 2D radio map estimation, this task focuses on the vertical dimension to capture spatial-spectral dependencies influenced by complex urban structures.

The proposed \textit{RadioFormer3D} model accepts two distinct input forms.
The first one is the sample information set $S$. 
Let $S = \{\mathbf{s}_1, \mathbf{s}_2, \dots, \mathbf{s}_k\}$ denote a set of $k$ sparse observations, where each sample $\mathbf{s}_i = (\mathbf{p}_i, v_i)$ consists of a 3D spatial coordinate $\mathbf{p}_i = (x_i, y_i, z_i)$ and its corresponding spectrum power value $v_i \in \mathbb{R}$.
The second input is the building-height map $\mathbf{M}$. To account for environmental factors, we introduce a building-height map $\mathbf{M} \in \mathbb{R}^{H \times W}$, where each pixel value indicates the building height at that specific horizontal location.

The goal is to learn a mapping function $\mathcal{F}$ such that
\begin{equation}
    \mathcal{F}(S, \mathbf{M}) = \hat{\mathbf{R}},
\end{equation}
where $\hat{\mathbf{R}} \in \mathbb{R}^{N \times H \times W}$ denotes the predicted 3D radio map with $N$ horizontal layers.
Given the limitations of existing datasets for producing continuous altitude maps, we define the task as a weakly-supervised learning problem. 
Specifically, only a limited set of discrete radio maps $\mathbf{R}_s \in \mathbb{R}^{N_s \times H \times W}$ is available as ground truth during training, where $N_s < N$. 
The model must generalize from these sparse labels to reconstruct a continuous, high-fidelity volumetric spectral field.

\subsection{Multi-Granularity Generative Architecture}
As illustrated in Fig.~\ref{fig_main}, the \textit{RadioFormer3D} architecture addresses the 3D radio map estimation task through three primary stages: multi-stream feature encoding via \textit{Dual-stream Self-Attention}~(DSA), cross-granularity feature fusion via \textit{Cross-stream Cross-Attention}~(CSCA), and volumetric reconstruction via a \textit{Multi-Height Decoder}.

The process begins with the DSA module, which captures multi-granularity features by independently processing heterogeneous inputs. Building upon the foundational \textit{RadioFormer}~\cite{radioformer}, we extend the DSA to handle spatial dynamics through two streams. 
In the patch-level stream, the building-height map $\mathbf{M}$ provides the fundamental environmental context. 
We employ a ViT~\cite{vit} architecture, denoted as $\varphi_v$, to extract global environmental features $\mathbf{f}_m = \varphi_v(\mathbf{M})$. 
By partitioning the map into patches, the model captures long-range spatial dependencies such as building clusters and street canyons.

To complement the environmental context, the pixel-level stream processes sparse 3D observations using an improved Point Encoder $\varphi_p$. Unlike traditional encoders, $\varphi_p$ transforms discrete samples $S = \{(\mathbf{p}_i, v_i)\}$ into dense point-level features through a coordinate-based modulation mechanism. To perceive high-frequency spatial variations, we first apply a Fourier positional encoding~\cite{fourier} to the 3D coordinates $\mathbf{p}_i$. 
For each 3D coordinate, the Fourier feature $\mathbf{f}_i$ is computed across $K$ frequency scales:
\begin{equation}
    \mathbf{f}_i = \operatorname{Concat}_{k=0}^{K-1} \left[ \sin(2^k \pi \mathbf{p}_i), \cos(2^k \pi \mathbf{p}_i) \right],
\end{equation}
where the sine and cosine functions are applied element-wise to $\mathbf{p}_i$, resulting in a $6K$-dimensional descriptor.
This representation is mapped to a geometric token $\mathbf{h}_i \in \mathbb{R}^D$ via an MLP~\cite{mlp}: $\mathbf{h}_i = \mathrm{MLP}_{\mathrm{coord}}(\mathbf{f}_i)$. Furthermore, we introduce an Adaptive Scale Modeling Process, where an adaptive radius $r_i$ representing the localized influence range is dynamically predicted:
\begin{equation}
    r_i = \mathrm{softplus}(\mathrm{MLP}_{r}([\mathbf{p}_i, v_i])) + \epsilon,
\end{equation}
where $\epsilon$ ensures positivity. To fuse signal intensity $v_i$ and scale $r_i$ into the geometric token $\mathbf{h}_i$, we adopt a residual-based Feature-wise Linear Modulation~(FiLM) formulation:
\begin{equation}
    \mathbf{z}_i = (1 + \alpha \cdot \tanh(\gamma_i)) \odot \mathbf{h}_i + \beta_i,
\end{equation}
where $\gamma_i$ and $\beta_i$ are generated from a condition vector $\mathbf{c}_i = [\mathbf{p}_i, v_i, r_i]$. The final modulated tokens $\mathbf{f}_p = \{\mathbf{z}_1, \dots, \mathbf{z}_k\}$ represent signal-aware 3D spatial features.

To integrate these heterogeneous streams, the CSCA module $\varphi_c$ facilitates cross-modal interaction by projecting the dense building features $\mathbf{f}_m$ and sparse signal features $\mathbf{f}_p$ into a unified latent space. Specifically, we compute the queries $\mathbf{Q}$, keys $\mathbf{K}$, and values $\mathbf{V}$ as follows:
\begin{equation}
    \mathbf{Q} = \mathbf{f}_m \mathbf{W}_Q, \quad \mathbf{K} = \mathbf{f}_p \mathbf{W}_K, \quad \mathbf{V} = \mathbf{f}_p \mathbf{W}_V,
\end{equation}
where $\mathbf{W}_Q, \mathbf{W}_K, \mathbf{W}_V \in \mathbb{R}^{d \times d}$ are learnable weight matrices. These linear transformations are essential for aligning the semantic representations of the geometric environment and the sparse radio observations, enabling the attention mechanism to effectively capture the spatial correlations between building obstructions and signal attenuation.
This mechanism enables the model to dynamically modulate signal propagation characteristics based on the surrounding environmental geometry:
\begin{equation}
    \mathbf{f}_c = \mathrm{Softmax}\left(\frac{\mathbf{Q} \mathbf{K}^\top}{\sqrt{d_k}}\right) \mathbf{V},
\end{equation}
where $d_k$ is the scaling factor. The resulting fused representation $\mathbf{f}_c$ captures a comprehensive spatial-spectral embedding.

Finally, the \textit{Multi-Height Decoder} projects the fused features into a dense volumetric map through a hierarchical process of feature mapping, stage blocks, and an output head. 
Initially, a $3\times3$ convolutional feature mapping module, denoted as $\varphi_f$, maps the latent features into the initial channel dimension, followed by residual blocks and a Non-Local self-attention module to enhance global dependency modeling. The network then enters a progressive decoding phase composed of multiple stage blocks $\varphi_s$. 
Each block stacks residual layers for feature transformation, with Non-Local modules inserted at specific low-resolution levels to enhance long-range awareness.
Starting from the second stage, upsampling operations double the spatial resolution while reducing the number of channels to facilitate coarse-to-fine reconstruction. To produce the final output, an output head $\varphi_o$ processes the features through GroupNorm and Swish activation, followed by a final $3\times3$ convolution to reconstruct the 3D radio map:
\begin{equation}
    \hat{\mathbf{R}} = \varphi_o \big( \varphi_s(\varphi_f(\mathbf{f}_c)) \big) \in \mathbb{R}^{N \times H \times W},
\end{equation}
where $\hat{\mathbf{R}}$ represents the predicted volumetric spectrum state, achieving final high-fidelity 3D radio map reconstruction.

\subsection{Joint Spectrum Integrity Loss}
To address the challenges of supervision scarcity and extreme spatial sparsity, we propose a multi-level loss strategy, the \textit{Joint Spectrum Integrity Loss} $\mathcal{L}$. 
This loss function $\mathcal{L}$ is composed of three main components: the \textit{Linear Volume Loss} $\mathcal{L}_{\mathrm{v}}$, the \textit{Radio Rendering Loss} $\mathcal{L}_{\mathrm{r}}$, and the \textit{Self-Sample Pixel Loss} $\mathcal{L}_{\mathrm{p}}$, providing supervision at the volume, map, and point levels, respectively. 
Throughout this work, we employ the Huber loss~\cite{Huber1992} as a robust regression objective to handle potential outliers in radio measurements. 
Unlike the Mean Squared Error $\mathcal{L}_{\mathrm{MSE}}$, which is highly sensitive to extreme values, the Huber loss provides a quadratic penalty for small errors and a linear penalty for large ones. 
This characteristic ensures that the model remains resilient to measurement noise and anomalous signal fluctuations.
For a scalar prediction $\hat{y}$ and ground truth $y$, the Huber loss is defined as:
\begin{equation}
\mathcal{L}_{\mathrm{h}}(\hat{y}, y)=
\begin{cases}
\frac{1}{2}(\hat{y}-y)^2, & \text{if } |\hat{y}-y|\le \delta,\\
\delta\left(|\hat{y}-y|-\frac{1}{2}\delta\right), & \text{otherwise},
\end{cases}
\end{equation}
where $\delta$ is a threshold parameter. For vector-valued prediction $\hat{\mathbf{y}}$ and ground truth $\mathbf{y}$, the Huber loss is applied element-wise and averaged over all dimensions $D$:
\begin{equation}
\mathcal{L}_{\mathrm{h}}(\hat{\mathbf{y}}, \mathbf{y})
=
\frac{1}{D}\sum_{i=1}^{D}\mathcal{L}_{\mathrm{h}}(\hat{y}_i, y_i).
\end{equation}

\begin{figure}[!t]
\centering
\includegraphics[width=0.49\textwidth]{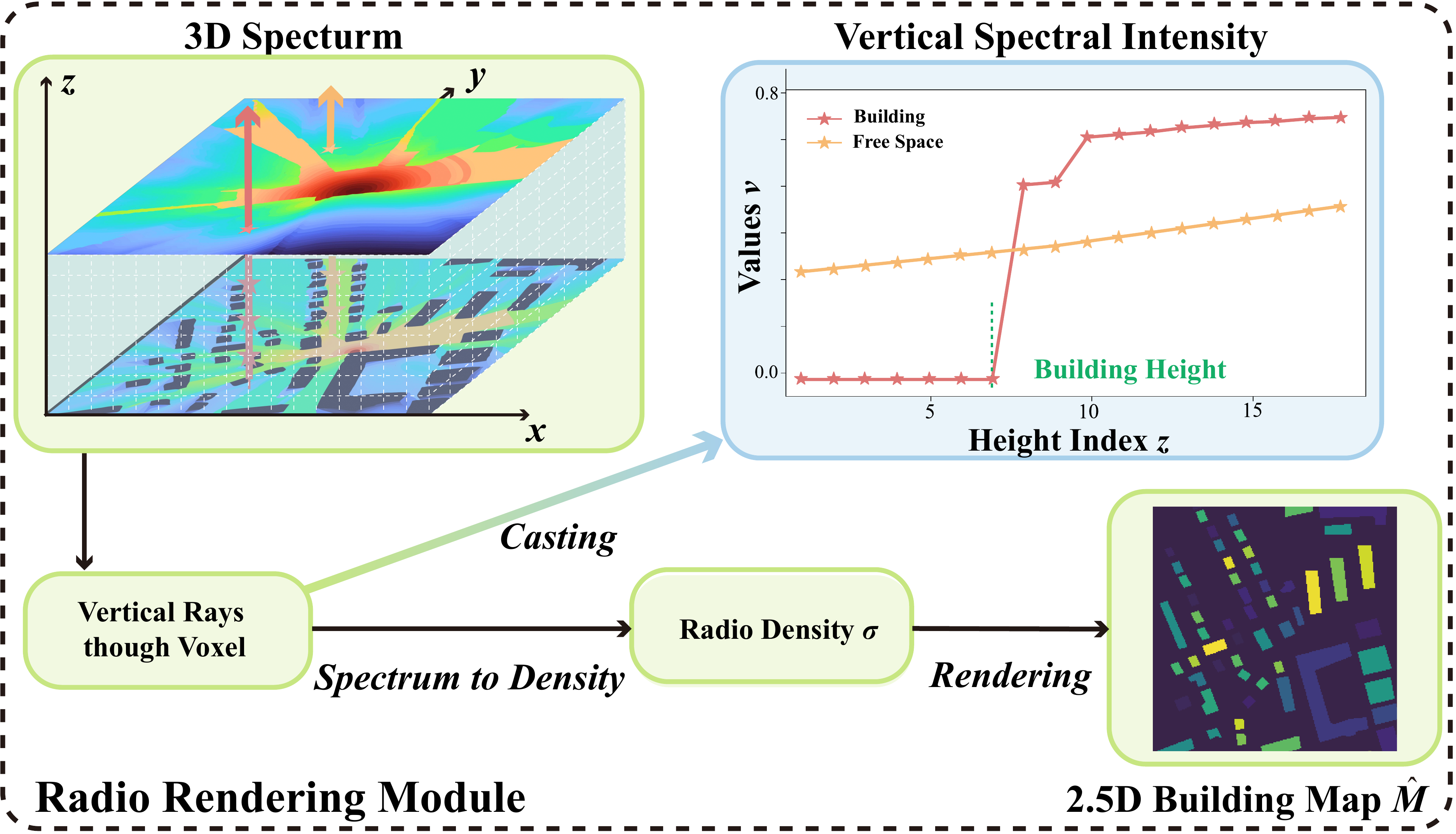}
\caption{
Overview of the RRM. The module casts vertical rays through the predicted 3D spectrum volume and analyzes the spectral intensity profile along the height dimension. By identifying the vertical transition between free space and building regions, the sampled intensities are converted into radio density $\sigma$, from which a 2.5D building-height map $\hat{\mathbf{M}}$ is rendered. In this way, the RRM injects structural environmental priors to regularize the learned volumetric spectrum field.
}
\label{fig_rrm}
\end{figure}
Based on this robust regression objective, we next formulate the proposed \textit{Joint Spectrum Integrity Loss}. 
First, to bridge the gap between the sparse supervision $\mathbf{R}_s \in \mathbb{R}^{N_s \times H \times W}$ and the dense volumetric prediction $\hat{\mathbf{R}} \in \mathbb{R}^{N \times H \times W}$, we introduce the \textit{Linear Volume Loss} $\mathcal{L}_{\mathrm{v}}$. 
The \textit{Linear Volume Loss} $\mathcal{L}_{\mathrm{v}}$ consists of two complementary parts. 
The first term follows the conventional supervised learning paradigm in RME by enforcing direct supervision on the observed altitude layers. 
The second term introduces a pseudo-label regularization strategy for the unsupervised layers. 
Inspired by Eq.~\ref{eq:log}, which characterizes radio propagation with an approximately linear attenuation trend, we construct dense pseudo-labels by linear interpolation along the vertical dimension.
Specifically, to extend the supervision from the observed altitude layers to the entire volume, we define a dense pseudo-label volume $\mathbf{G}_l \in \mathbb{R}^{N \times H \times W}$ through piecewise linear interpolation along the altitude dimension. 
For any target altitude $z \in \{0,\dots,N-1\}$, the pseudo-label slice $\mathbf{G}_l(z)$ is defined as:
\begin{equation}
\mathbf{G}_l(z)=
\begin{cases}
\mathbf{R}_s^{(1)}, & \text{if } z \le z_1,\\
\mathbf{R}_s^{(N_s)}, & \text{if } z \ge z_{N_s},\\
\mathbf{R}_s^{(k)}+\dfrac{z-z_k}{z_{k+1}-z_k}\left(\mathbf{R}_s^{(k+1)}-\mathbf{R}_s^{(k)}\right), & \text{otherwise},
\end{cases}
\end{equation}
where $z_k$ and $z_{k+1}$ are the two adjacent supervised altitudes surrounding $z$.
Based on $\mathbf{G}_l$, the $\mathcal{L}_{\mathrm{v}}$ is formulated as:
\begin{equation}
\mathcal{L}_{\mathrm{v}}=
\frac{1}{N_s}\sum_{k=1}^{N_s}\mathcal{L}_{\mathrm{h}}\!\left(\hat{\mathbf{R}}^{(z_k)}, \mathbf{R}_s^{(k)}\right)
+\lambda_{pl}\,\mathcal{L}_{\mathrm{h}}\!\left(\hat{\mathbf{R}}, \mathbf{G}_l\right),
\end{equation}
where $z_k$ denotes the $k$-th supervised altitude, $\mathbf{R}_s^{(k)}$ is the corresponding observed spectrum slice, and $\lambda_{pl}$ is a weighting coefficient. In this way, $\mathcal{L}_{\mathrm{v}}$ not only preserves fidelity at the observed layers but also propagates supervision to unsampled altitudes, thereby encouraging vertically consistent reconstruction throughout the full 3D spectrum volume.

To provide structural regularization in the vertical dimension, we further introduce the \textit{Radio Rendering Loss} $\mathcal{L}_{\mathrm{r}}$ through a specialized Radio Rendering Module~(RRM). 
As illustrated in Fig.~\ref{fig_rrm}, the RRM casts vertical rays through the predicted 3D spectrum volume and extracts the spectral intensity values $\{v_i\}_{i=1}^{N}$ along each voxel column. 
The resulting vertical spectral profiles reveal a clear difference between free-space and building regions: the free-space response varies smoothly with height, while the building response exhibits a pronounced structural change near the building boundary. 
Therefore, the vertical transition point naturally provides a cue for estimating the building height. 
Based on this observation, the RRM converts the sampled spectral responses into radio densities and performs differentiable rendering along the height dimension, yielding a 2.5D building-height map $\hat{\mathbf{M}}$.
Specifically, to decouple signal intensity from geometric density while preserving their structural correlation, we define the radio density at altitude $z_i$ as:
\begin{equation}
\sigma_i = \operatorname{sigmoid}(k(v_i - t)),
\end{equation}
where $k$ and $t$ are learnable parameters. The rendering process is then formulated by the opacity $\alpha_i$ and accumulated transmittance $T_i$:
\begin{equation}
\alpha_i = 1 - \exp(-\sigma_i \Delta z_i), \quad
T_i = \prod_{j=1}^{i-1}(1-\alpha_j),
\end{equation}
where $\Delta z_i = z_{i+1}-z_i$ denotes the interval between adjacent altitude samples. 
Based on these quantities, the rendered height is computed as:
\begin{equation}
\hat{\mathbf{M}}=\sum_{i=1}^{N} w_i z_i, \quad w_i=T_i\alpha_i.
\end{equation}
Accordingly, the $\mathcal{L}_{\mathrm{r}}$ could be defined as:
\begin{equation}
\mathcal{L}_{\mathrm{r}} = \mathcal{L}_{\mathrm{h}}(\hat{\mathbf{M}}, \mathbf{M}),
\end{equation}
where $\mathbf{M}$ denotes the ground-truth building-height map. In this way, $\mathcal{L}_{\mathrm{r}}$ injects environmental structural priors into the 3D spectrum reconstruction and improves vertical consistency by explicitly constraining the rendered building-height structure.

Complementary to the volumetric and structural constraints, the \textit{Self-Sample Pixel Loss} $\mathcal{L}_{\mathrm{p}}$ provides fine-grained supervision at the original sparse sampling locations $S$. 
This loss is designed to preserve both the local statistical characteristics and the global distribution of the observed spectrum samples. 
Specifically, the first term aligns the mean and variance of the predicted and observed values within each altitude bin, while the second term enforces consistency between their overall marginal distributions.
To this end, the sampling set $S$ is first partitioned into $K$ disjoint altitude bins $\{\mathcal{B}_1,\dots,\mathcal{B}_K\}$ according to the height coordinate $z_i$. 
For each bin, we compute the mean $\mu_k$ and standard deviation $\sigma_k$ of the observed spectrum power values $v_i$, while $\hat{\mu}_k$ and $\hat{\sigma}_k$ denote the corresponding statistics of the predicted values $\hat{v}(\mathbf{p}_i)$. 
In addition, let $\mathbf{P}$ and $\mathbf{Q}$ denote the normalized soft-binned histograms of the predicted and observed spectrum power values over $S$, respectively, where soft binning is implemented using a Gaussian kernel. 
Let $\mathbf{E}=\frac{1}{2}(\mathbf{\mathbf{P}}+\mathbf{Q})$ denote their mixture distribution. The pixel loss is then formulated as:
\begin{equation}
\begin{aligned}
\mathcal{L}_{\mathrm{p}}
=&\sum_{k=1}^{K}\left[\mathcal{L}_{\mathrm{h}}(\hat{\mu}_k,\mu_k)+\mathcal{L}_{\mathrm{h}}(\hat{\sigma}_k,\sigma_k)\right] \\
&+\frac{1}{2}\mathcal{D}_{\mathrm{KL}}(\mathbf{P} \,\|\, \mathbf{E})
+\frac{1}{2}\mathcal{D}_{\mathrm{KL}}(\mathbf{Q} \,\|\, \mathbf{E}).
\end{aligned}
\end{equation}
In this way, $\mathcal{L}_{\mathrm{p}}$ jointly preserves the local statistical structure and the global distributional consistency of the reconstructed spectrum field at the sparse sampling locations.

Finally, the overall \textit{Joint Spectrum Integrity Loss} training objective is formally defined as:
\begin{equation}
    \mathcal{L} = \lambda_{\mathrm{v}} \mathcal{L}_{\mathrm{v}} + \lambda_{\mathrm{r}} \mathcal{L}_{\mathrm{r}} + \lambda_{\mathrm{p}} \mathcal{L}_{\mathrm{p}},
\end{equation}
where $\lambda_{\mathrm{v}}$, $\lambda_{\mathrm{r}}$, and $\lambda_{\mathrm{p}}$ are hyperparameters that balance volumetric continuity, structural alignment, and point-level distribution consistency.

\section{Experiments}
\label{exp}

\begin{table*}[t]
\centering
\caption{Main experimental results on UrbanRadio3D and SpectrumNet datasets. We compare six models across three settings: (1) Supervision with MSE Loss $\mathcal{L}_{MSE}$, (2) Weak Supervision without the proposed Joint Spectrum Integrity Loss $\mathcal{L}$, and (3) Weak Supervision with $\mathcal{L}$. The \underline{\textbf{bold and underlined}} values indicate the best performance, while the \underline{\textit{italicized and underlined}} values represent the second-best performance.}
\label{tab:main_results}
\small
\newcolumntype{Y}{>{\centering\arraybackslash}X}
\begin{tabularx}{\textwidth}{@{} l | YYY | YYY | YYY @{}}
\toprule
\multirow{2}{*}{\textbf{Method}} & \multicolumn{3}{c|}{\textbf{Supervision~(with $\mathcal{L}_\mathrm{MSE}$)}} & \multicolumn{3}{c|}{\textbf{Weak Supervision (with $\mathcal{L}_\mathrm{MSE}$)}} & \multicolumn{3}{c}{\textbf{Weak Supervision (with $\mathcal{L}$)}} \\ \cmidrule(l){2-10} 
 & RMSE$\downarrow$ & PSNR$\uparrow$ & SSIM$\uparrow$ & RMSE$\downarrow$ & PSNR$\uparrow$ & SSIM$\uparrow$ & RMSE$\downarrow$ & PSNR$\uparrow$ & SSIM$\uparrow$ \\ \midrule
\rowcolor[HTML]{F5F5F5} 
\multicolumn{10}{l}{\textbf{Dataset: \textit{UrbanRadio3D}~\cite{radiodiff3d}}} \\ \midrule
\textit{UNet}~\cite{unet} & 0.0545 & 25.3482 & 0.8976 & 0.2531 & 12.0774 & 0.1703 & 0.0861 & 21.3608 & 0.7855 \\
\textit{RadioUNet}~\cite{radiounet} & 0.0512 & 25.9912 & 0.8288 & 0.2716 & 11.9802 & 0.1632 & 0.0812 & 21.5133 & 0.7759 \\
\textit{PMNet}~\cite{pmnet} & 0.1076 & 23.6930 & 0.7875 & \underline{\textit{0.2391}} & \underline{\textit{12.1120}} & \underline{\textit{0.2012}} & 0.0969 & 20.3569 & 0.7526 \\
\textit{RadioDUN}~\cite{radiodun} & 0.0459 & 25.9124 & 0.8690 & 0.2431 & 11.9972 & 0.2981 & 0.0810 & 22.1254 & 0.7789 \\
\textit{DAT-UNet}~\cite{datunet} & \underline{\textbf{0.0330}} & \underline{\textbf{30.4003}} & \underline{\textbf{0.9389}} & 0.2624 & 11.6915 & 0.1810 & \underline{\textit{0.0752}} & \underline{\textit{22.5498}} & \underline{\textit{0.7865}} \\
\rowcolor[HTML]{FFFFC7} 
\textit{RadioFormer3D} & \underline{\textit{0.0366}} & \underline{\textit{29.2342}} & \underline{\textit{0.9089}} & \underline{\textbf{0.1764}} & \underline{\textbf{16.3926}} & \underline{\textbf{0.3241}} & \underline{\textbf{0.0730}} & \underline{\textbf{22.7913}} & \underline{\textbf{0.7827}} \\ \midrule
\rowcolor[HTML]{F5F5F5} 
\multicolumn{10}{l}{\textbf{Dataset: \textit{SpectrumNet}~\cite{spectrumnet}}} \\ \midrule
\textit{UNet}~\cite{unet} & 0.1421 & 0.4482 & 16.9577 & 0.2313 & 11.2324 & 0.2011 & 0.1399 & 0.4532 & 17.0452 \\
\textit{RadioUNet}~\cite{radiounet} & 0.1332 & 0.4895 & 14.5232 & 0.2213 & 11.2567 & 0.2199 & 0.1232 & 0.5022 & 15.1923 \\
\textit{PMNet}~\cite{pmnet} & 0.1108 & 0.5785 & 19.1248 & 0.2543 & 12.2134 & 0.1920 & 0.1001 & 0.5892 & 19.2319 \\
\textit{RadioDUN}~\cite{radiodun} & 0.1073 & 0.6387 & 20.1815 & \underline{\textit{0.1802}} & \underline{\textit{12.7830}} & \underline{\textit{0.2289}} & 0.0991 & 0.6092 & 19.3465 \\
\textit{DAT-UNet}~\cite{datunet} & \underline{\textit{0.0923}} & \underline{\textit{0.6451}} & \underline{\textit{20.3399}} & 0.1932 & 12.9231 & 0.2201 & \underline{\textit{0.0961}} & \underline{\textit{0.6321}} & \underline{\textit{19.9902}} \\
\rowcolor[HTML]{FFFFC7} 
\textit{RadioFormer3D} & \underline{\textbf{0.0838}} & \underline{\textbf{0.6983}} & \underline{\textbf{21.6563}} & \underline{\textbf{0.1709}} & \underline{\textbf{13.0012}} & \underline{\textbf{0.2341}} & \underline{\textbf{0.0922}} & \underline{\textbf{0.6512}} & \underline{\textbf{20.9912}} \\ \bottomrule
\end{tabularx}
\end{table*}

\subsection{Experiment Setup}
In this section, we provide a detailed description of the experimental configurations used to evaluate \textit{RadioFormer3D}, including the datasets, comparative methods, evaluation protocols, and implementation details.

\textbf{Datasets.}
To evaluate the proposed framework's performance, we conduct experiments on two large-scale datasets: \textit{UrbanRadio3D}~\cite{radiodiff3d} and \textit{SpectrumNet}~\cite{spectrumnet}.

\textit{UrbanRadio3D} is a high-resolution 3D radio map dataset designed for 6G environment-aware communications. \
Unlike traditional 2D datasets that focus on planar pathloss, \textit{UrbanRadio3D} captures complex electromagnetic propagation in realistic urban geometries through high-fidelity ray-tracing simulations. 
The dataset provides full 3D spatial coverage with 20 vertical altitude layers~(from 1m to 20m) and includes three key metrics: Pathloss, Direction of Arrival~(DoA), and Time of Arrival~(ToA). 
It contains over 2.84 million labeled data points with a fine-grained spatial resolution of $1\text{m}^3$ per voxel. 
The simulations use a center carrier frequency of 5.9 GHz, a fixed transmit power of 23 dBm/Hz, and isotropic antennas. 
Each radio map covers an area of $20 \times 256 \times 256 \text{ m}^3$. 
In our experiments, we focus on the Pathloss metric across 19 altitude layers (1m to 19m). 
To simulate weakly-supervised learning settings, we use only the 1m, 10m, and 19m layers for training and validation, while the remaining layers are reserved for evaluating the model's volumetric reconstruction performance.

\textit{SpectrumNet} is an open benchmark for 6G communications and generative AI research, featuring multi-dimensional data across three spatial dimensions and five frequency bands. 
It provides vertical information at altitudes of 1.5m, 30m, and 200m, covering signal distributions from ground level to low-altitude airspace. 
To avoid the influence of inconsistent propagation characteristics across different frequency bands, we select data from the 1.5 GHz and 1.7 GHz bands for our experiments. 
The dataset ensures physical accuracy by accounting for building occlusions, terrain elevation, and atmospheric absorption, in accordance with ITU-R P.676 and ITU-R P.838-3. 
Since it provides radio maps at only three altitudes, for our setup, we use the 1.5m and 200m layers for training and validation, respectively, and reserve the intermediate 30m layer for volumetric reconstruction during testing.

For both datasets, we partition the data into training, validation, and testing sets with ratios of 0.7:0.1:0.2. 
To prevent potential data leakage and ensure the model's generalization, we ensure that the geographic terrains and urban layouts in the testing set are not present in the training or validation sets. 
This strategy ensures the model evaluates its performance in unseen environments rather than memorizing building geometry.

\textbf{Evaluation Metrics.} To quantitatively evaluate the performance of our proposed framework, we adopt three widely used metrics. 
First, the Root Mean Square Error~(RMSE) $\downarrow$ measures the average pixel-wise intensity difference between the predicted and ground-truth radio maps. 
Second, the Peak Signal-to-Noise Ratio~(PSNR) $\uparrow$ is utilized to quantify reconstruction fidelity by comparing the maximum possible signal power to the power of corrupting noise. 
Finally, we employ the Structural Similarity Index~(SSIM)~\cite{ssim} $\uparrow$ to assess the perceptual quality of the generated maps. 
SSIM provides a comprehensive evaluation by considering luminance, contrast, and structural information, thereby better reflecting the spatial consistency of the electromagnetic environment. 

\begin{figure}[!t]
\centering
\includegraphics[width=\columnwidth]{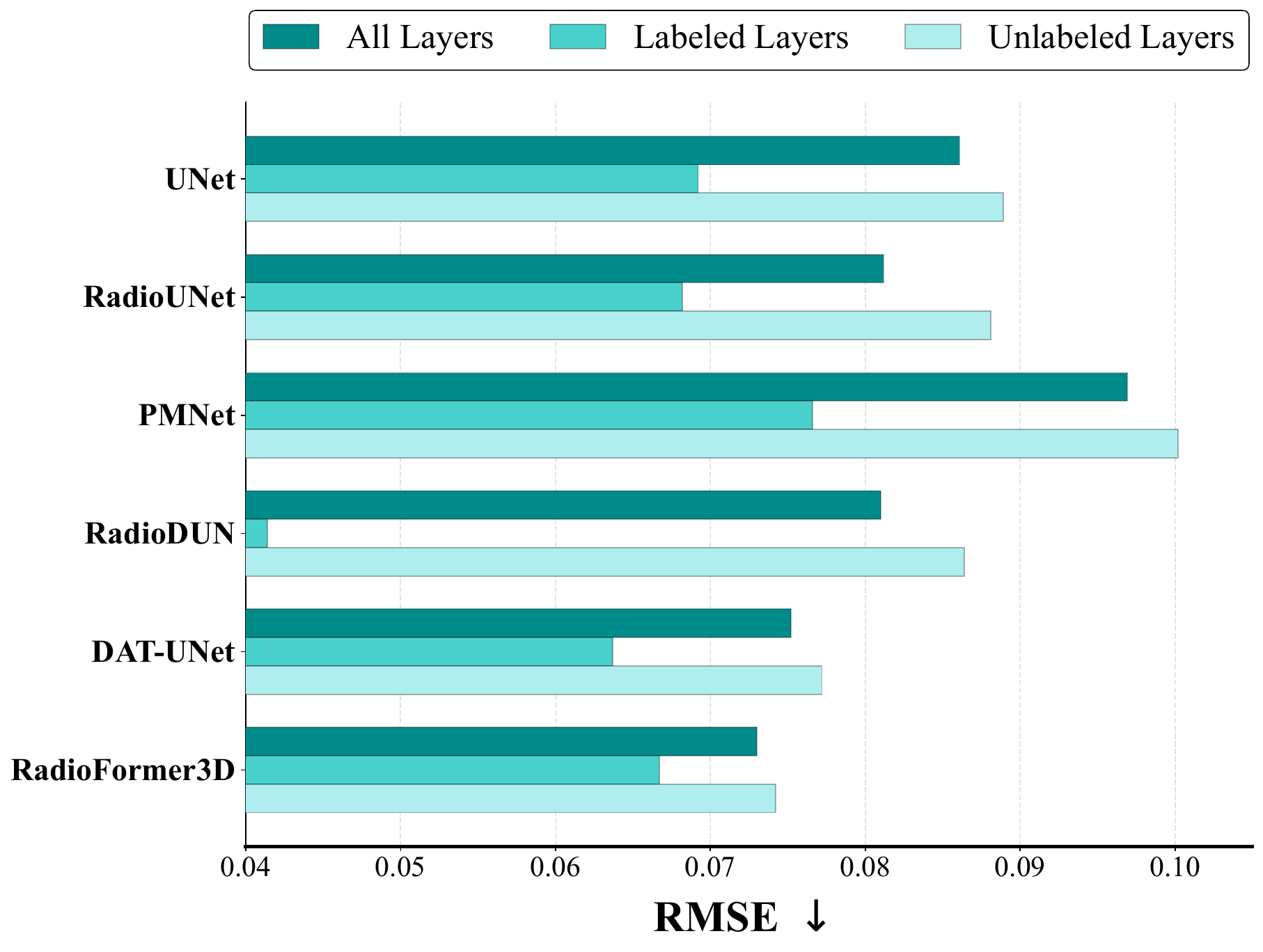}
\caption{Comparison of reconstruction error~(RMSE) across different altitude layers. 
We categorize the 3D volume into labeled layers~(where sparse supervision is available) and unlabeled layers (reconstructed via model inference). 
}
\label{fig_5}
\end{figure}

\textbf{Comparative Methods}
To evaluate the performance of \textit{RadioFormer3D}, we compare it against several state-of-the-art~(SOTA) generative architectures specifically designed for or widely applied to radio map estimation. 
These baselines include: \textit{UNet}~\cite{unet}, a fundamental encoder-decoder benchmark for spatial field reconstruction; 
\textit{RadioUNet}~\cite{radiounet}, which explicitly models shadowing and diffraction by incorporating building geometry; 
\textit{PMNet}~\cite{pmnet}, which integrates physical heuristics for enhanced pathloss prediction; 
\textit{RadioDUN}~\cite{radiodun}, a physics-driven deep unfolding model for spectrum situation generation; and \textit{DAT-UNet}~\cite{datunet}, which leverages Deformable Attention Transformers to capture long-range spatial dependencies. 
Notably, while diffusion-based models such as \textit{RadioDiff}~\cite{radiodiff} and \textit{RMDM}~\cite{rmdm} have shown promise in radio map generation, they are ill-suited for our 3D weakly-supervised setting; the extreme vertical sparsity of available ground truth prevents these models from establishing stable generative priors through iterative denoising. Consequently, we focus our evaluation on the aforementioned RME models, utilizing their official open-source implementations and default configurations to ensure a fair and reproducible comparison.

\begin{figure*}[t]
\centering
\includegraphics[width=\textwidth]{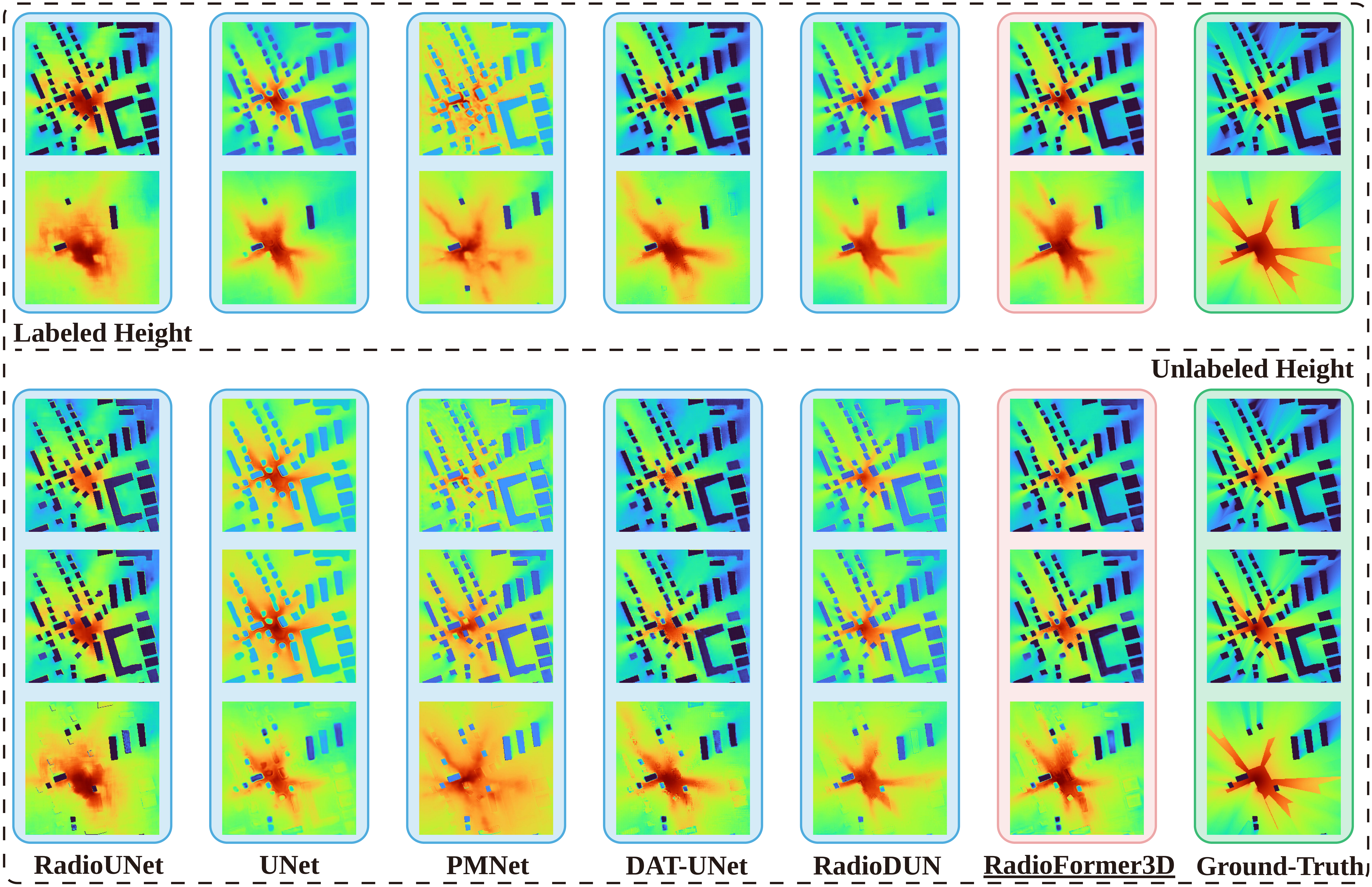}
\caption{Visualization of 3D radio map reconstruction on the UrbanRadio3D dataset. The rows are categorized into Labeled Heights~(1m and 19m), used during the training phase, and Unlabeled Heights~(3m, 7m, and 17m), which represent unseen layers for evaluating weakly supervised performance. 
Each column displays the reconstruction results of our proposed \underline{RadioFormer3D} and several baseline models~(\textit{RadioUNet}, \textit{UNet}, \textit{PMNet}, \textit{DAT-UNet}, and \textit{RadioDUN}), along with the corresponding ground-truth.}
\label{fig_main_vis}
\end{figure*}

\textbf{Implementation Details.}
To ensure a fair and rigorous comparison, all experiments are conducted under identical settings unless otherwise specified. 
The number of spatial sample points is fixed at 50, so the input to all comparative models consists of 50 sparse sampling points and a building-height map. 
All models are optimized using AdamW with a cosine annealing learning rate scheduler. 
The initial learning rate is set to $1\times10^{-3}$ and the weight decay is $1\times10^{-4}$. 
Training is performed for 100 epochs on a server equipped with eight NVIDIA GeForce RTX 4090 GPUs, and validation is conducted every 10 epochs to monitor performance and prevent overfitting. 
Under this hardware configuration, \textit{RadioFormer3D} requires approximately 2 minutes per epoch for training. 
To mitigate stochasticity, each model is evaluated five times, and the average performance is reported.

For the loss configuration, the balancing weights are fixed across all experiments, with $\lambda_{\mathrm{v}}=1.0$, $\lambda_{\mathrm{r}}=0.05$, and $\lambda_{\mathrm{p}}=0.1$.  
The Huber threshold is set to $\Delta z_i=0.1$. 
For the RRM, the vertical step size $\delta_i$ is set to 1.0 for \textit{UrbanRadio3D} and 10.0 for \textit{SpectrumNet}. 
For the radio rendering loss term, the coefficient $\lambda_{pl}$ is set to 0.3.
For the point-level supervision branch, the number of altitude bins in $\mathcal{L}_{\mathrm{m}}$ is set to 8, and the number of soft histogram bins in $\mathcal{L}_{\mathrm{his}}$ is set to 2.

\subsection{Comparison With State-of-the-Art Methods}

\textbf{Quantitative Comparison and Performance Robustness.} 
Table~\ref{tab:main_results} reports the quantitative results of \textit{RadioFormer3D} and five representative baselines under three settings: full supervision, weak supervision without the proposed loss $\mathcal{L}$, and weak supervision with $\mathcal{L}$. 
Under full supervision, high-capacity models such as \textit{DAT-UNet} achieve strong fitting performance. 
However, when supervision becomes sparse, their performance degrades substantially, indicating limited robustness under weak supervision. 
A notable observation is that introducing the proposed \textit{Joint Spectrum Integrity Loss} $\mathcal{L}$ consistently improves the performance of all compared methods in the weakly-supervised setting, demonstrating that $\mathcal{L}$ serves as an effective and architecture-agnostic regularizer rather than benefiting only the proposed model. 
Among all methods, \textit{RadioFormer3D} attains the best results in the weakly-supervised setting with $\mathcal{L}$, achieving an RMSE of 0.0730 on \textit{UrbanRadio3D} and 0.0922 on \textit{SpectrumNet}. 
A direct comparison between the two weakly-supervised settings further highlights the effectiveness of $\mathcal{L}$. Specifically, on \textit{UrbanRadio3D}, adding $\mathcal{L}$ improves the RMSE/PSNR of \textit{RadioFormer3D} from 0.1764/16.3926 to 0.0730/22.7913. 
On \textit{SpectrumNet}, the RMSE is reduced from 0.1709 to 0.0922, with consistent gains in PSNR and SSIM. 
These results demonstrate that $\mathcal{L}$ provides effective regularization under sparse supervision by constraining the reconstruction from volumetric, structural, and point-distribution perspectives, thereby significantly enhancing the robustness and generalization ability of \textit{RadioFormer3D}.

\begin{table}[t]
\centering
\caption{Efficiency comparison on the UrbanRadio3D dataset. The table evaluates reconstruction accuracy against computational complexity (Parameters, FLOPs, and Inference Time). Our proposed RadioFormer3D is highlighted in gray.}
\label{tab:efficiency}
\newcolumntype{Y}{>{\centering\arraybackslash}X}
\begin{tabularx}{\columnwidth}{@{}l|Y Y Y Y@{}}
\toprule
\textbf{Method} & \textbf{RMSE} $\downarrow$ & \textbf{Params} & \textbf{FLOPs} & \textbf{Inference} \\ \midrule
\textit{UNet}          & 0.0861 & 0.24M  & 6.00G & 0.017s \\
\textit{RadioUNet}     & 0.0812 & 13.39M & 9.99G & 0.034s \\
\textit{PMNet}         & 0.0969 & 35.25M & 51.91G & 0.031s \\
\textit{RadioDUN}      & 0.0810 & 27.42M & 24.50G & 0.066s \\
\textit{DAT-UNet}      & 0.0752 & 29.35M & 10.12G & 0.065s \\
\rowcolor[HTML]{EFEFEF} 
\textbf{\textit{RadioFormer3D}} & \textbf{0.0730} & \textbf{5.39M} & \textbf{9.19G} & \textbf{0.018s} \\ \bottomrule
\end{tabularx}
\end{table}

\textbf{Qualitative Analysis and Physical Fidelity.} 
The visual evidence in Fig.~\ref{fig_main_vis} provides deeper insight into the models' inferential capabilities.
At unlabeled altitudes~(3m, 7m, 17m), baseline models such as \textit{UNet} and \textit{PMNet} produce ``structural artifacts'' and blurred boundaries, indicating a failure to maintain spatial continuity in the absence of vertical supervision. 
Conversely, \textit{RadioFormer3D} reconstructs sharp shadowing effects and fine-grained diffraction patterns that closely align with the ground truth. 
This high-fidelity reconstruction at unobserved altitudes implies that our model has transitioned from simple image-to-image mapping to capturing the intrinsic physical logic of 3D wave propagation, effectively recovering missing information through learned environmental priors.

\textbf{Vertical Generalization.} 
The decomposition of RMSE across altitude layers, as shown in Fig.~\ref{fig_5}, reveals a fundamental limitation of existing volumetric reconstruction methods. Most baselines suffer from a vertical generalization gap, characterized by a sharp escalation in error when transitioning from labeled to unlabeled layers. For instance, while \textit{RadioDUN} achieves high precision in labeled layers~(RMSE=0.0414), its error nearly doubles in unlabeled regions~(RMSE=0.0864), suggesting that conventional models often settle for discrete representations rather than learning a truly integrated 3D manifold.
In contrast, \textit{RadioFormer3D} maintains a remarkably stable error profile across the vertical dimension. The marginal performance gap between its labeled~(RMSE=0.0667) and unlabeled~(RMSE=0.0742) layers demonstrates that our model effectively captures cross-altitude spatial correlations. By leveraging the continuous nature of the 3D radio field via the Radio Rendering Module, \textit{RadioFormer3D} bridges the supervision gap, generalizing beyond local samples to reconstruct a consistent, physically informed 3D spectrum volume.

\textbf{Computational Efficiency.} 
The performance comparison regarding parameter count, inference latency, and reconstruction accuracy is summarized in Table~\ref{tab:efficiency}.
\textit{RadioFormer3D} overcomes this limitation by achieving SOTA accuracy with a compact footprint of 5.3858M parameters. By optimizing the attention mechanism for volumetric radio data, our model achieves an inference speed of 0.0182s, surpassing complex unfold-based models such as \textit{RadioDUN} by $3.5\times$. This Pareto-optimal balance between memory footprint, inference latency, and reconstruction precision establishes \textit{RadioFormer3D} as a highly scalable solution for real-world 3D RME construction.

\subsection{Ablation Study}
In this section, we conduct a series of ablation experiments on the UrbanRadio3D dataset to evaluate the individual contribution and effectiveness of each component within our proposed Joint Spectrum Integrity Loss ($\mathcal{L}$).

\begin{table}[t]
\centering
\caption{Ablation study on UrbanRadio3D dataset. FPE ($\varphi_v$) denotes the Fourier Point Encoder. $\mathcal{L}_v, \mathcal{L}_p, \mathcal{L}_r$ represent Linear Volume, Self-Sample Pixel, and Radio Rendering losses, respectively.}
\label{tab:ablation1}
\resizebox{\columnwidth}{!}{
\begin{tabular}{@{}l|ccc@{}}
\toprule
\textbf{Configuration} & \textbf{RMSE} $\downarrow$ & \textbf{PSNR} $\uparrow$ & \textbf{SSIM} $\uparrow$ \\ \midrule
Baseline (Transformer) & 0.1984 & 15.9112 & 0.5122 \\
+ FPE ($\varphi_v$) & 0.1764 & 16.3926 & 0.5601 \\ \midrule
\textit{Linear Pseudo Label (Ref.)} & \textit{0.0792} & \textit{22.0328} & \textit{0.7211} \\ \midrule
+ $\varphi_v$ + $\mathcal{L}_v$ & 0.0791 & 22.2365 & 0.7210 \\
+ $\varphi_v$ + $\mathcal{L}_v$ + $\mathcal{L}_p$ & 0.0748 & 22.6346 & 0.7712 \\
+ $\varphi_v$ + $\mathcal{L}_v$ + $\mathcal{L}_r$ & 0.0779 & 22.2764 & 0.7655 \\
+ Joint Spectrum Integrity Loss $\mathcal{L}$ & \textbf{0.0730} & \textbf{22.7913} & \textbf{0.7827} \\ \bottomrule
\end{tabular}%
}
\end{table}

\textbf{Analysis of Component Contributions.}
As illustrated in Table \ref{tab:ablation1}, we systematically evaluate performance by incrementally integrating the proposed loss terms. 
The results demonstrate that each component is vital for enhancing 3D spectrum generation. 
Specifically, the inclusion of the Linear Volume Loss $\mathcal{L}_v$ establishes a baseline for volumetric continuity and significantly reduces global reconstruction errors in unsampled regions. The Radio Rendering Loss $\mathcal{L}_r$ further improves structural alignment by enforcing physical consistency with the 2.5D building environment. Finally, the Self-Sample Pixel Loss $\mathcal{L}_p$ refines the fine-grained signal distribution through moment and histogram alignment, yielding the highest fidelity in both the statistical and spatial domains. The consistent performance gains observed from integrating each module validate the necessity of this multidimensional supervisory strategy. Beyond quantitative improvements, our experimental analysis reveals an intriguing phenomenon: the numerical discrepancy between the generated linear pseudo-labels $\mathbf{g}_l$ and the actual ground truth is significantly larger than the final inference loss achieved by \textit{RadioFormer3D}. This finding is noteworthy because it suggests that the model does not merely replicate the simplistic piecewise-linear interpolation provided by the pseudo-labels. Instead, by synergizing sparse sampling information with environmental geometric features, \textit{RadioFormer3D} demonstrates a robust potential to learn the underlying physical laws of radio propagation. In effect, the model acts as a physically informed refiner that generalizes beyond the imperfect supervision of pseudo-labels to reconstruct more accurate and realistic 3D spectral fields.

\begin{table}[t]
\centering
\caption{Impact of supervision altitude distribution on UrbanRadio3D dataset. $N_s$ denotes the number of supervised layers, and $\{z\}$ represents the specific altitudes (meters).}
\label{tab:ablation_altitude}
\begin{tabularx}{\columnwidth}{@{} c | l | X X X @{}}
\toprule
$N_s$ & \textbf{Supervised Altitudes} $\{z\}$ & \textbf{RMSE} $\downarrow$ & \textbf{PSNR} $\uparrow$ & \textbf{SSIM} $\uparrow$ \\ \midrule
\multirow{3}{*}{1} & $\{1\}$ & 0.1770 & 15.1036 & 0.6636 \\
 & $\{10\}$ & 0.1168 & 18.7939 & 0.7314 \\
 & $\{19\}$ & 0.1319 & 17.6469 & 0.6825 \\ \midrule
\multirow{2}{*}{2} & $\{1, 19\}$ & 0.0839 & 21.6412 & 0.7410 \\
 & $\{10, 11\}$ & 0.1109 & 19.2215 & 0.7505 \\ \midrule
\multirow{2}{*}{3} & \textbf{$\{1, 10, 19\}$} & 0.0730 & 22.7913 & 0.7827 \\
 & $\{9, 10, 11\}$ & 0.1089 & 19.3907 & 0.7602 \\ \midrule
5 & $\{1, 6, 10, 15, 19\}$ & 0.0692 & 23.3267 & 0.8100 \\
7 & $\{2, 5, 8, \dots, 19\}$ & 0.0619 & 24.2710 & 0.8187 \\
9 & $\{2, 4, 6, \dots, 18\}$ & 0.0592 & 24.7054 & 0.8463 \\ \midrule
19 & \textit{All (Full Supervision)} & 0.0366 & 29.2342 & 0.9089 \\ \bottomrule
\end{tabularx}
\end{table}

\begin{figure}[t]
\centering
\includegraphics[width=\columnwidth]{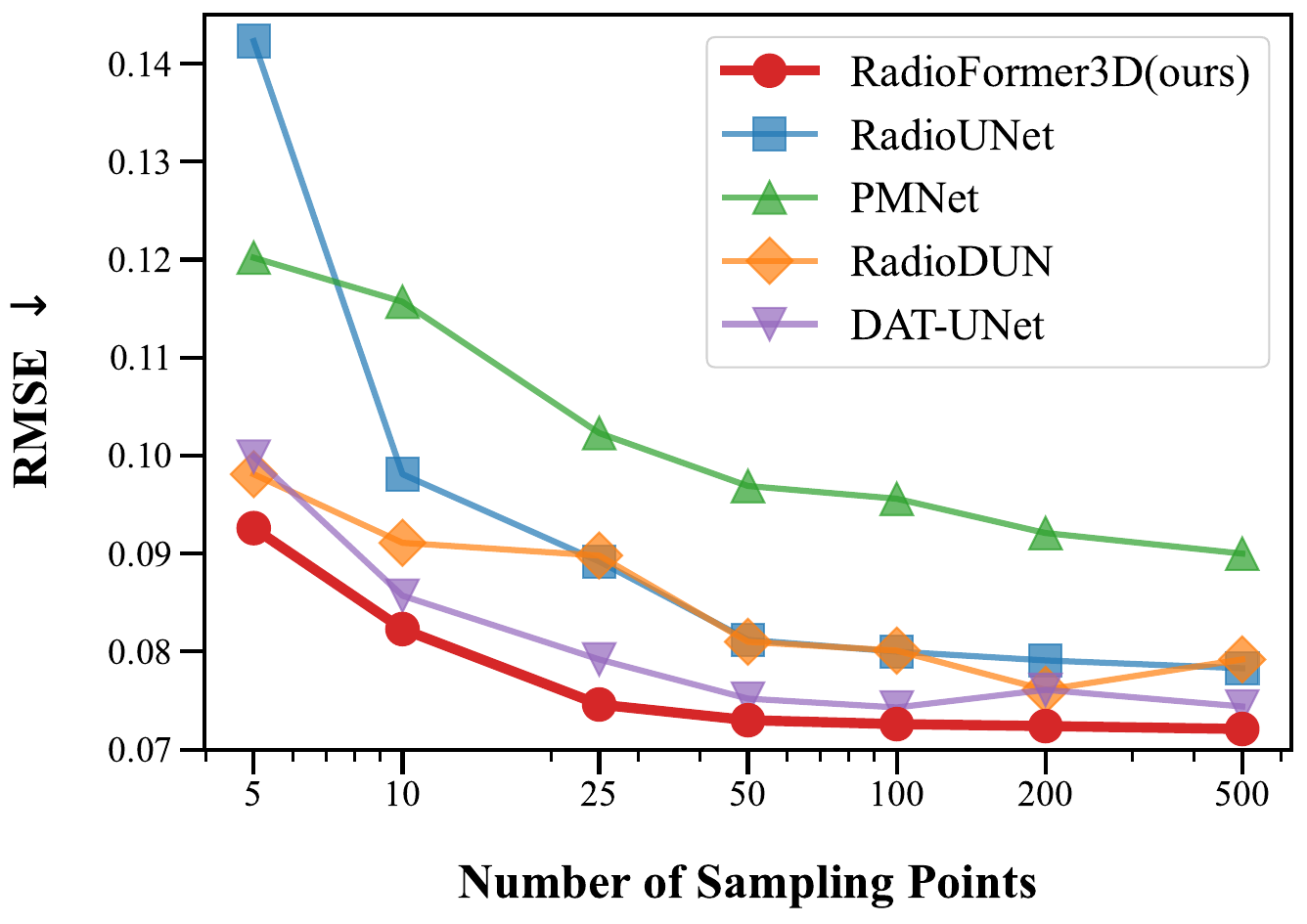}
\caption{Impact of supervision density on reconstruction performance. The figure compares RMSE performance across varying numbers of sparse observation samples, specifically $S_n\in \{5, 10, 25, 50, 100, 200, 500\}$.}
\label{fig_sampling_density}
\end{figure}

\textbf{Impact of Supervision Altitude Strategy.} 
Table~\ref{tab:ablation_altitude} presents the performance of \textit{RadioFormer3D} under varying supervision altitude configurations, revealing several critical insights into vertical propagation modeling. 
First, model accuracy improves monotonically with the number of supervised layers $N_s$, as evidenced by the RMSE dropping from 0.1770 with a single layer to 0.0366 with full supervision. 
Second, vertical diversity proves more critical than proximity; a significant performance gap exists between clustered and spread-out supervision. 
For instance, with $N_s=3$, the distributed configuration $\{1, 10, 19\}$m outperforms the clustered $\{9, 10, 11\}$m strategy by approximately 32\% in RMSE, suggesting that supervision at the volume boundaries and median better captures vertical gradients and propagation variations. 
Third, a centrality bias is observed in single-layer settings, where supervision at the median altitude ($\{10\}$m) yields superior results compared to the ground ($\{1\}$m) or top ($\{19\}$m) layers, indicating that the middle layer provides a more representative for the 3D space. 

\textbf{Sensitivity to Sampling Point Density.} 
To evaluate the robustness of our framework under extremely sparse supervision, we analyze the performance of \textit{RadioFormer3D} and several state-of-the-art baselines across a range of sampling densities, from 5 to 500 points per layer. As illustrated in Fig.~\ref{fig_sampling_density}, our model consistently outperforms all baselines across the entire density spectrum, exhibiting superior data efficiency. In the regime of extreme sparsity, such as with only 5 or 10 sampling points, baselines like \textit{RadioUNet} suffer from severe performance degradation with RMSE exceeding 0.14, whereas \textit{RadioFormer3D} maintains a remarkably low error of 0.0926, matching the performance of competing models that utilize 10 to 20 times more data. Furthermore, the performance of \textit{RadioFormer3D} converges rapidly; beyond 50 sampling points, the marginal gain in reconstruction accuracy diminishes, indicating that the Joint Spectrum Integrity Loss effectively captures the underlying physical laws even with minimal point-wise supervision. This efficiency is primarily due to the RRM and the Linear Volume Loss $\mathcal{L}_v$, which together transform the task from purely data-driven point estimation to physically constrained volumetric reconstruction. By leveraging environmental geometry as a structural prior, \textit{RadioFormer3D} bridges the gap between discrete sparse samples and continuous 3D spectrum fields, demonstrating robust generalization capabilities essential for practical large-scale radio map deployments.

\section{Conclusion}
\label{conclusion}
In this paper, we propose \textit{RadioFormer3D}, an estimation model for the 3D RME task under weak supervision. 
At the architectural level, \textit{RadioFormer3D} introduces several targeted extensions for volumetric reconstruction. 
Specifically, a Fourier-based sampling encoder maps sparse 3D measurements into a high-dimensional space to improve geometric representation, and a specialized Multi-Height Decoder reconstructs dense radio distributions across continuous altitudes. 
By introducing the \textit{Joint Spectrum Integrity Loss}, our approach effectively reduces the vertical generalization gap in 3D environments and addresses the issue of sparse supervision along the height dimension. This design enables \textit{RadioFormer3D} to learn wave propagation patterns across different altitudes by combining structural constraints with multi-level statistical alignment. 
Experimental results on multiple benchmarks show that \textit{RadioFormer3D} achieves state-of-the-art performance, preserving high structural fidelity at both observed and unobserved altitudes while maintaining practical efficiency. Although 3D reconstruction remains challenging due to the larger solution space and the simplified treatment of urban diffraction effects, this work provides a strong baseline for weakly supervised volumetric estimation. 
We believe \textit{RadioFormer3D} is a useful step toward high-fidelity 3D electromagnetic characterization for emerging 6G systems and low-altitude economy scenarios.

\bibliographystyle{IEEEtran}
\bibliography{IEEEabrv, contents/cite}
\end{document}